\documentclass{article}

\usepackage{microtype}
\usepackage{graphicx}
\usepackage{subfigure}
\usepackage{booktabs} 
\usepackage{multirow}
\usepackage{float}

\usepackage{hyperref}

\usepackage[accepted]{icml2024}

\usepackage{amsmath}
\usepackage{amssymb}
\usepackage{mathtools}
\usepackage{amsthm}

\usepackage[capitalize,noabbrev]{cleveref}

\theoremstyle{plain}

\theoremstyle{definition}

\theoremstyle{remark}

\usepackage{listings}
\usepackage{color}

\definecolor{codegreen}{rgb}{0,0.6,0}

\lstdefinelanguage{diff}{ 
  morecomment=[f][\color{blue}]{"""C},
  morecomment=[f][\color{blue}]{\ \ \ \ """I}, 
  morecomment=[f][\color{blue}]{import},  
  morecomment=[f][\color{blue}]{def},  
  morecomment=[f][\color{blue}]{\ \ \ \ """F},     
  morecomment=[f][\color{blue}]{\ \ \ \ return},
  morecomment=[f][\color{codegreen}]{\#},
  morecomment=[f][\color{red}]{"""A},
}

\lstdefinelanguage{diff2}{
  morecomment=[f][\color{codegreen}]{\ \ \ \ ('sc},
  morecomment=[f][\color{codegreen}]{\ \ \ \ ('mo},
  morecomment=[f][\color{codegreen}]{\ \ \ \ \ \ \ \ s},
  morecomment=[f][\color{red}]{"""A},
}

\usepackage[textsize=tiny]{todonotes}

\icmltitlerunning{Entropy-Regularized Token-Level Policy Optimization}

\begin{document}

\twocolumn[
\icmltitle{
Entropy-Regularized Token-Level Policy Optimization \\
for Language Agent Reinforcement
}



\icmlsetsymbol{equal}{*}

\begin{icmlauthorlist}
\icmlauthor{Muning Wen}{aff1}
\icmlauthor{Junwei Liao}{aff3}
\icmlauthor{Cheng Deng}{aff1}
\icmlauthor{Jun Wang}{aff2}
\icmlauthor{Weinan Zhang}{aff1}
\icmlauthor{Ying Wen}{aff1}
\end{icmlauthorlist}

\icmlaffiliation{aff1}{Shanghai Jiao Tong University}
\icmlaffiliation{aff2}{University College London}
\icmlaffiliation{aff3}{Xi'an Jiaotong  University}

\icmlcorrespondingauthor{Ying Wen}{ying.wen@sjtu.edu.cn}

\icmlkeywords{Machine Learning, ICML}

\vskip 0.3in
]



\printAffiliationsAndNotice{}  

\begin{abstract}
Large Language Models (LLMs) have shown promise as intelligent agents in interactive decision-making tasks. Recent approaches often depend on meticulously designed prompts, high-quality examples, or additional reward models for in-context learning, supervised fine-tuning, or RLHF. Reinforcement learning (RL) presents a dynamic alternative for LLMs to overcome these dependencies by engaging directly with task-specific environments. Nonetheless, it faces significant hurdles: 1) instability stemming from the exponentially vast action space requiring exploration; 2) challenges in assigning token-level credit based on action-level reward signals, resulting in discord between maximizing rewards and accurately modeling corpus data. In response to these challenges, we introduce Entropy-Regularized Token-level Policy Optimization (ETPO), an entropy-augmented RL method tailored for optimizing LLMs at the token level. At the heart of ETPO is our novel per-token soft Bellman update, harmonizing RL process with the principles of language modeling. It decomposes the Q-function update from a coarse action-level view to a more granular token-level perspective, backed by theoretical proof of optimization consistency. Crucially, this decomposition renders linear time complexity in action exploration. We assess the effectiveness of ETPO within a simulated environment that models data science code generation as a series of multi-step interactive tasks; results show that ETPO achieves effective performance improvement on the CodeLlama-7B model and surpasses a variant PPO baseline inherited from RLHF. This underlines ETPO's potential as a robust method for refining the interactive decision-making capabilities of LLMs\footnote{Our code is open-sourced at \url{https://github.com/morning9393/ETPO}}. For a more detailed preliminary work describing our motivation for token-level decomposition and applying it in PPO methods, please refer to \citet{wen2024reinforcing}.

\end{abstract}

\section{Introduction}

The capability of large language models (LLMs) to function as intelligent agents has been showcased across various domains, such as mathematical research~\cite{romera2023mathematical}, data science~\cite{hollmann2023large}, software engineering~\cite{qian2023communicative}, and geoscience~\cite{Deng2023K2AF,Lin2023GeoGalacticaAS}. Given these developments, there is an increasing interest in enhancing their functionality in interactive decision-making contexts~\cite{shridhar2020alfworld}.
Historically, the enhancement of LLMs has relied heavily on the crafting of precise prompts and the provision of high-quality examples for in-context learning or supervised fine-tuning (SFT)~\cite{yao2022react, shinn2023reflexion, radford2018improving}.
Fortunately, employing reinforcement learning (RL)~\cite{sutton2018reinforcement} offers a dynamic approach for LLMs to actively participate and progressively refine their performance in task-specific environments through a process of exploration and feedback. This shift towards RL allows for a reduction in dependency on manually prepared examples and prompts, paving the way for more autonomous and adaptive learning capabilities in LLMs.

Although RL is inherently designed for tackling problems in interactive environments, its direct application~\cite{carta2023grounding, ramamurthy2022reinforcement, ouyang2022training} for fine-tuning language models encounters two significant gaps. The first pertains to a pronounced misalignment in learning objectives. Specifically, while the goal of language modeling is to approximate the probability distribution of tokens in the corpus, RL focuses on the maximization of quantified rewards (which in return has optimal state-action distribution, a.k.a. occupancy measure or policy, with maximized long-term discounted return).
Such a focus, however, predisposes the model to overfitting risks~\cite{zhang2018study}. This fundamental divergence undermines the preservation of the original token distribution in language models, a critical aspect for maintaining linguistic coherence and diversity~\cite{ziegler2019fine}.

The emergence of a second gap is attributed to the variations in optimization granularity. RL typically focuses on optimizing at the action level, whereas language modeling targets optimization at the token level. This discrepancy leads to two primary issues: 1) The action space for LLMs, consisting of sequences of tokens, is exponentially vast, leading to increased instability and convergence difficulties during the RL fine-tuning process~\cite{shinn2023reflexion}. 2) The reward signals at the action level are overly sparse for sequences of tokens, complicating the provision of precise, fine-grained supervision for complex tasks. This results in obstacles in assigning credit to individual tokens and misalignment between token generation and action performance~\cite{carta2023grounding, chen2024improving}. 
For instance, RLHF~\cite{ziegler2019fine} seeks to bridge the first gap by integrating a KL divergence term within the reward signal. However, the effectiveness of this approach is heavily dependent on an additional reward model that delivers dense token-level reward signals. Despite this, applying RLHF with action-level rewards still encounters the aforementioned difficulties, which are elaborately analyzed in Section\ref{sub_sec_convergence}. Moreover, RLHF conceptualizes token sequences as temporal relations, which hinders its utility in multi-step interactive settings. This limitation stems from a confusion between assigning credit within a single action and across multiple actions.

In this paper, we take several steps to bridge these gaps between RL and language modeling. 
Initially, to mitigate the risk of the LLM policy diverging excessively during RL fine-tuning, we incorporate interactive tasks within the entropy-regularized reinforcement learning (ERL) framework, which reinterprets the RL problems through the lens of probability theory and constrains the policy update with a given distribution~\cite{schulman2017equivalence, levine2018reinforcement}. Thus we can naturally bridge the first gap in learning objectives, i.e. aligning the demand for improving task-specific performance with the objectives of language modeling. Then, to address the second gap concerning optimization granularity, we propose the \textit{per-token soft Bellman update} and \textit{per-token policy update}. They decompose the Q function~\cite{sutton2018reinforcement} and policy optimization within ERL from the action level to the token level, backed by theoretical proof affirming their consistency in optimization. This decomposition reduces the complexity growth of action space exploration as context length from multiplicative to additive, thus stabilizing the learning process. Besides, it provides fine-grained credit assignment, further ensuring that each token generation is directly correlated with action performance. Unlike RLHF, our method distinctly maintains the separation between intra-action and inter-action credit assignments during token-level policy updates, enhancing its effectiveness in multi-step interactions.

As a natural outcome of our findings, we propose \textit{Entropy-regularized Token-level Policy Optimization (ETPO)}, a concrete RL algorithm that leverages the per-token soft Bellman update and policy update to enhance LLM agents' capabilities within interactive environments. Our experiments confirm the effectiveness of ETPO in narrowing the gaps in both learning objectives and optimization granularity. We justify such a claim by evaluating ETPO in an environment that simulates data science code generation as multi-step interactive tasks; results show that ETPO achieves effective performance improvement on the CodeLlama-7B model~\cite{touvron2023llama} and surpasses a variant PPO~\cite{schulman2017proximal} baseline inherited from RLHF. 

\begin{figure}[!ht]
\vskip -0.1in
\label{fig_code_iter}
    \centering
\begin{lstlisting}[language=diff2,linewidth=  {\linewidth},frame=tb,basicstyle=\footnotesize\ttfamily]
# Environment Step=1, Reward=0.7463
model = RandomForestClassifier()

# Environment Step=16, Reward=0.8716
model = Pipeline([
    ('scaler', StandardScaler()), 
    ('logistic_regression', LogisticRegression())])

# Environment Step=224, Reward=0.8968
pipeline = Pipeline([
    ('scaler', StandardScaler()),
    ('model', MLPClassifier())])

# Environment Step=496, Reward=0.9289
pipeline = Pipeline([
    ('scaler', StandardScaler()),
    ('model', MLPClassifier(
        solver='lbfgs', batch_size=100,
        ......))])
\end{lstlisting}
    \vskip -0.15in
    \caption{Exemplary run of ETPO on the Balance Scale dataset. This figure illustrates the evolution of code generated by CodeLlama-7B during the experiment where the environment step means the number of interactions. We've highlighted the positive changes that resulted in performance improvements in each iteration, e.g. finding better models or hyper-parameters. This investigation implies that ETPO can guide models for the emergence of complex yet more effective behaviors.}
\vskip -0.1in
\end{figure}

\section{Related Works}
\subsection{LLMs for Interactive Tasks}
In many works like Self-Refine~\cite{madaan2023self}, ReAct, Reflexion, or ProTeGi~\cite{pryzant2023automatic}, they "verbally" tell the model how to iteratively improve the quality of what they generated. Reasoning via Planning (RAP)~\cite{hao2023reasoning} combines the Monte Carlo Tree Search (MCTS) with LLMs' self-evaluation mechanism, to balance the exploration and exploitation in the vast reasoning space. Self-Rewarding~\cite{yuan2024self} allows LLMs to prompt themselves to provide rewards during training. 
These works excel in numerous tasks, yet their success hinges on the foundational models possessing robust inherent capabilities~\cite{feng2023alphazero}. 

RL4LMs~\cite{ramamurthy2022reinforcement} build an open-source framework for optimizing language generators interactively with RL algorithms. CodeRL~\cite{le2022coderl} generates code as actions and optimizes in an RL framework using an actor-critic setup to debug programs given feedback from the runtime environment. To align the objective of token-level optimization with action-level optimization, GLAM~\cite{carta2023grounding} estimates the probability of possible actions with the conditional probability of tokens composing the actions, and updates the action as a whole with the PPO algorithm. 
Nevertheless, the effectiveness of these reinforcement learning-based approaches is constrained by the excessively large action space encountered during exploration and the lack of fine-grained supervision~\cite{shinn2023reflexion, chen2024improving}.

\subsection{Entropy-Regularized Reinforcement Learning}
\label{subsec_erl}
j methods~\cite{schulman2017equivalence, ziebart2010modeling, fox2015taming, haarnoja2017reinforcement, nachum2017bridging} are a subset of reinforcement learning that aims to search for the optimal stochastic policy under the constraints of a given distribution. ERL constrains the learned policy from deviating too far from a reference distribution by adding an entropy-regularized term to the reward function and optimizing with soft Bellman update~\cite{schulman2017equivalence}. A concrete instance of ERL is the Soft Actor-Critic~\cite{haarnoja2018soft, haarnoja2018soft_applications} algorithm, which defines the $\bar{\pi}$ to be a uniform distribution to balance the trade-off between exploration (entropy maximization) and exploitation (reward maximization). While PPO and some other Actor-Critic methods~\cite{konda1999actor} also encourage exploration by adding an entropy term as a penalty to the loss function, this penalty only takes effect in the current time step. RLHF~\cite{ziegler2019fine} that adopts a variant of PPO with a KL term added to its reward function could also be seen as an approximation to ERL.

\section{Preliminaries}

\subsection{Language-Augmented Sequential Decision-Making}
\label{sub_sec_formulation}
Sequential decision-making problems with linguistic inputs and outputs can be framed as a language-augmented Markov Decision Process (MDP)~\cite{van2012reinforcement} $\mathcal{M}=(\mathcal{V}, \mathcal{S}, \mathcal{A}, \mathcal{T}, r, \gamma)$\footnote{For notation convenience, we omit the partial observability and thus the observation functions that take a global state as an input and outputs a local observation.}. Given a vocabulary $\mathcal{V}$ and tokens $w\in\mathcal{V}$, $\mathcal{S}\subset\mathcal{V}^N$, $\mathcal{A}\subset\mathcal{V}^N$ are the state space and action space respectively, which all consist of sequences of tokens. $\mathcal{T}: \mathcal{S}\times\mathcal{A}\mapsto\mathcal{S}$ is the state transition function. $r: \mathcal{S}\times\mathcal{A}\mapsto\mathbb{R}$ is the reward function and $\gamma$ is the discounted factor. 

When we introduce an LLM as policy $\pi$, at time step $t\in\mathbb{N}$, $\pi$ receives a textual state $s_t\in\mathcal{S}$ from the environment as input and generates an action $a_t\in\mathcal{A}$. The action $a_t$ is normally a sentence or phrase that consists of a sequence of token $a_t=(w_t^1,\dots,w_t^{|a_t|})$ and generated in an auto-regressive manner token-by-token. Then, the generated textual action $a_t$ will be mapped to the specific API call or command and executed in an interactive environment~\cite{schick2023toolformer}. After execution, the environment returns a reward signal $r(s_t, a_t)$ along with the next state $s_{t+1}$, according to the transition $\mathcal{T}$. 

\subsection{Entropy-Regularized Reinforcement Learning}
\label{subsec_ps_2_pi}
One gap between RL and language modeling is that they hold different learning objectives. While LLMs aim to model the probability distribution of tokens within a corpus~\cite{koller2009probabilistic}, traditional RL focuses on maximizing the discounted cumulative return: 
\begin{align}
\mathcal{G}'(\pi)=\mathbb{E}_{a\sim\pi}\Big[\sum_{t=0}^{\infty}\gamma^t r(s_t, a_t)\Big],
\end{align}
which is prone to overfitting~\cite{zhang2018study}. 
Directly applying RL to LLMs poses significant risks of disrupting the well-calibrated token distribution of LLMs and thus impairs the diversity and readability of language expressions that are essential for language models.

Fortunately, there have been many studies~\cite{todorov2008general, kappen2012optimal, kappen2011optimal, ziebart2010modeling, toussaint2009robot, levine2018reinforcement} trying to couch the RL problem in the parlance of probability theory and develop a set of ERL methods. The expected entropy-regularized return which ERL targets to maximize is:
\begin{align}
\label{equ_erl_objective}
\mathcal{G}(\pi)=\mathbb{E}_{a\sim\pi}\Big[\sum_{t=0}^{\infty}\gamma^t(r(s_t, a_t)-\beta D_{KL}[\pi||\bar{\pi}](s_t))\Big],
\end{align}
where $D_{KL}[\pi||\bar{\pi}](s_t)=D_{KL}[\pi(\cdot|s_t)||\bar{\pi}(\cdot|s_t)]$ is the KL divergence between the learning policy $\pi$ and reference policy $\bar{\pi}$, $\beta$ is a scalar coefficient to balance the influence between reward signals and KL divergence. The corresponding soft Bellman update for the Q function is:
\begin{align}
\label{equ_soft_bellman}
Q(s_t,a_t) \leftarrow & ~r(s_t,a_t)+\gamma(\mathbb{E}_{a_{t+1}\sim\pi}[Q(s_{t+1},a_{t+1})] \nonumber\\
& - \beta D_{KL}[\pi||\bar{\pi}](s_t)).
\end{align}
Equation~\ref{equ_erl_objective} indicates that the objective of ERL is constrained with a given probability distribution $\bar{\pi}$. Thus, by assigning the $\bar{\pi}$ with an original LLM $\rho$ in Equation~\ref{equ_soft_bellman}, we can smoothly bridge the gap in learning objectives between RL and language modeling, i.e. maximizing the cumulative return while maintaining the original token distribution as much as possible.



\subsection{Action-Level and Token-Level Policy Optimization}
\label{subsec_ao_2_to}
Another gap between RL and language modeling is the differences in the granularity of optimization.
In most traditional reinforcement learning scenarios with discrete action spaces, actions are typically inseparable, thus rendering the action space $\mathcal{|A'|} = O(d)$, where $d$ is the number of available actions. However, as described in Section~\ref{sub_sec_formulation}, when we introduce LLMs as policies, an action $a$ consists of a sequence of tokens $a=(w^1,\dots,w^{|a|})$ generated by the LLM. In this case, since each token has an independent search space with vocabulary size $O(|\mathcal{V}|)$, the action space will experience an exponential explosion as the length of the context increases: $\mathcal{|A|}=O(|\mathcal{V}|^l)$, where $l$ is the context length. Thus, directly optimizing for the entire action shown in Equation~\ref{equ_glam} needs to explore an enormous action space, e.g. $|\mathcal{V}|=32016$ and $l=256$ in our experimental setting, and thus suffer from instability and difficulty in converging during RL fine-tuning.
\begin{align}
\label{equ_glam}
    \pi(a|s) = \prod_{j=1}^n \pi\left(w^j \mid w^1, w^2, \ldots, w^{j-1}\right).
\end{align}
Besides, a reward signal is obtained after executing a complete action,  which is too sparse to provide fine-grained supervision for each token. Applying it to all tokens within an action as Equation~\ref{equ_share_credit} might lead to a misalignment between token generation and action performance, where each token holds different importance in expressing the overall meaning and effectiveness of an action~\cite{chen2024improving}. 
\begin{align}
\label{equ_share_credit}
Q(s_t,w_t^{1:j-1},w_t^j) \leftarrow Q(s_t, a_t), ~\forall j=1,\dots,|a_t|.
\end{align} 
For instance, while a language model-generated code yielding positive returns doesn't imply that each line is beneficial, discarding certain useless or harmful segments could enhance overall outcomes, provided the algorithm identifies them during optimization. 


Inspired by the decomposition for high-dimensional continuous actions or multi-agent joint actions~\cite{chebotar2023q, wen2022multi}, we can consider sequentially decomposing the optimization from the action level into the token level as well. While reducing the growth of action space as the length of token sequence from $O(|\mathcal{V}|^l)$ to $O(|\mathcal{V}|\times l)$ and enabling fine-grained credit assignment, we should ensure the consistency in the optimization process before and after decomposition.


\section{Entropy-Regularized Token-Level Policy Optimization}

\begin{figure*}[!ht]
\vskip -0.1in
\begin{center}
\centerline{\includegraphics[width=0.9\linewidth]{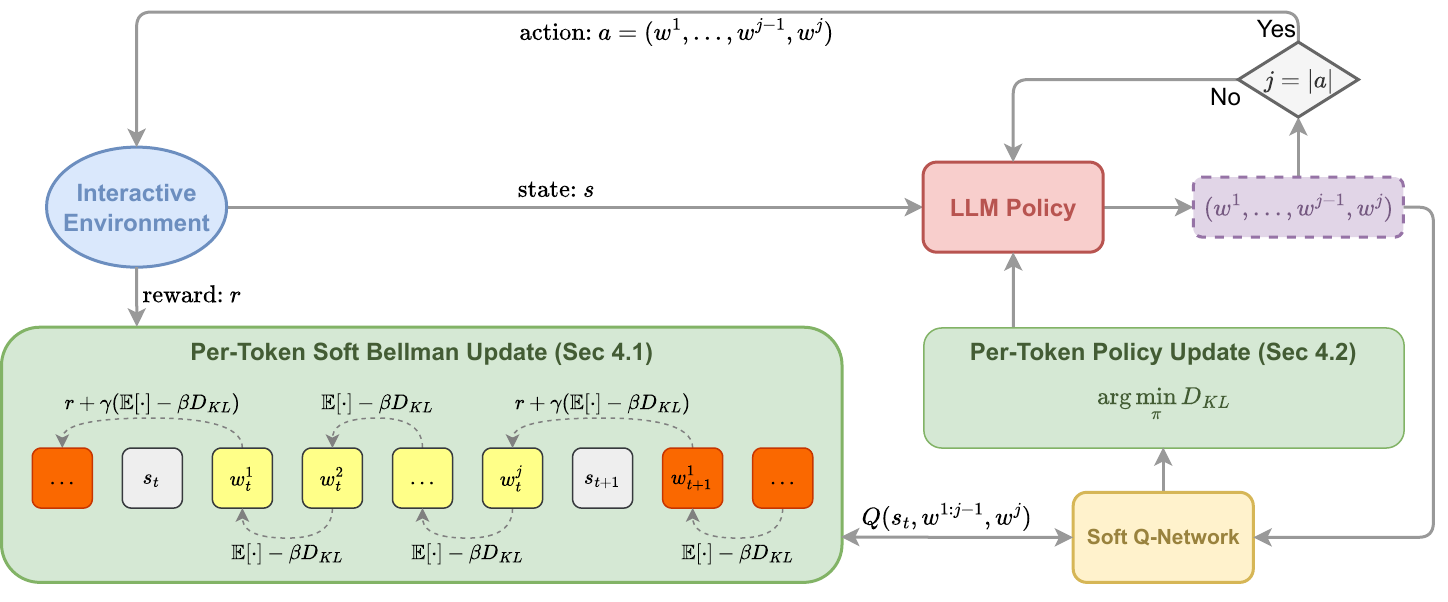}}
\vskip -0.1in
\caption{The overall pipeline of ETPO. In each time step, the LLM agent receives a state from the interactive environment. Then it generates an action token-by-token until the action is ready and being executed, i.e. $j=|a|$. Then the action will be separated into a sequence of tokens and their Q values will be updated following the per-token soft Bellman update scheme, where the $\mathbb{E}[\cdot]-\beta D_{KL}$ and $r+\gamma(\mathbb{E}[\cdot]-\beta D_{KL})$ are respectively corresponding to the different cases in Equation~\ref{equ_per_token_soft_bellman}. The LLM policy will also be updated toward minimizing the KL divergence between it and the soft Q-network. 
}
\label{fig_main} 
\vskip -0.3in
\end{center}
\end{figure*}

Based on the motivation discussed in Sections~\ref{subsec_ps_2_pi} and~\ref{subsec_ao_2_to}, we propose a concrete token-level ERL algorithm that yields the highest expected entropy-augmented return, i.e. the Equation~\ref{equ_erl_objective}. The intuition behind our method is to decompose a textual action into a sequence of tokens and backpropagate proper credit for each token with a variant of soft Bellman update, i.e. the per-token soft Bellman update in Equation~\ref{equ_per_token_soft_bellman}. That way, we can learn a token-level soft Q-function, bypassing the requirement to calculate a precise $\mathbb{E}_{a_{t+1}\sim\pi}[Q(s_{t+1}, a_{t+1})]$. After learning the soft Q-function, we can update policy $\pi$ by minimizing the KL divergence between it and the exponent of the learned soft Q-function, the per-token policy update in Equation~\ref{equ_per_token_policy_update}. Figure~\ref{fig_main} illustrates the workflow of our proposed method.

\begin{figure*}[!ht]
\centering
\vskip -0.1in
\begin{align}
\label{equ_per_token_soft_bellman}
Q(s_t,w_t^{1:j-1},w_t^j)\leftarrow
\begin{cases} 
\mathbb{E}_{w_t^{j+1}\sim\pi(s_t,w_t^{1:j})}[Q(s_t,w_t^{1:j},w_t^{j+1})]-\beta D_{KL}[\pi||\bar{\pi}](s_t,w_t^{1:j}),  & \mbox{if } j < |a_t| \\
r(s_t,a_t)+\gamma(\mathbb{E}_{w_{t+1}^1\sim\pi(s_{t+1})}[Q(s_{t+1},w_{t+1}^1)]-\beta D_{KL}[\pi||\bar{\pi}](s_{t+1})), & \mbox{if } j = |a_t|
\end{cases}
\end{align}
\vskip -0.2in
\centering
\begin{align}
\label{equ_per_token_policy_update}
\pi_{new} = \arg\min_{\pi}\mathcal{J}_{Q}(\pi) = \arg\min_{\pi}D_{KL}\Big[\pi(\cdot|s_t,w_t^{1:j-1})||\pi_Q^*(\cdot|s_t, w_t^{1:j-1})\Big].
\end{align}
\vskip -0.2in
\end{figure*}


\subsection{Per-Token Soft Bellman Updates}
\label{subsec_per_token_soft_bellman}

Decomposing an action into a sequence of tokens, $a=(w^1,\dots,w^{|a|})$, where $w\in\mathcal{V}$ is the tokens and $|a|$ is the length of the sequence, we extend the soft Bellman update in Equation~\ref{equ_soft_bellman} to a per-token view of actions, and define a per-token soft Bellman update in Equation~\ref{equ_per_token_soft_bellman}. For a time step $t$, $w_t^{1:j}$ denotes the sequence of tokens in action $a_t$ from the first token $w_t^1$ to the $j$-th token $w_t^j$, i.e. $w_t^{1:j}=(w_t^1,\dots,w_t^j)$. Then we define the soft Q-value of the $j$-th token $w_t^j$ conditioned on the textual state $s_t$ and previous action tokens $w_t^{1:j-1}$ in an auto-regressive manner, without discount when $j<|a_t|$ (first line in the equation). When $j=|a_t|$, The reward $r(s_t,a_t)$ is applied to the last token (second line in the equation), as we do not receive any reward before executing the whole action. Besides, we discount the expected Q-value and KL-divergence which are conditioned on the next state $s_{t+1}$ and the corresponding first token $w_{t+1}^1$. In short, we only discount Q-values between the time steps and keep the discounted factor at 1.0 for all but the last token within actions in each time step. 

After this decomposition, The most important thing is to make sure whether it will disturb the alignment between token generation and action selection. By decomposing each action into a sequence of tokens for the soft Bellman update following the Equation~\ref{equ_per_token_soft_bellman}, we do not change the general optimization properties and the principle of the Bellman optimality still holds for a given MDP. We theoretically maintain the consistency between the objective of token-level optimization and action-level optimization, where the proof is left in Appendix~\ref{appendix_soft_q_update}. 

With this per-token soft Bellman update, we can ensure the generation process for each token toward maximizing the expected cumulative return, that is, predicting the next token is directly incentivized to solve the decision-making problem in an interactive environment currently. Additionally, we can bypass the calculation for precise $\mathbb{E}_{a_{t+1}\sim\pi}[Q(s_{t+1}, a_{t+1})]$ that suffers from an unmanageable computational complexity of $O(d^N)$, instead, we just need to calculate $|a_{t+1}|$ token-level expected values $\mathbb{E}_{w_t^{j+1}\sim\pi(s_t,w_t^{1:j})}[Q(s_t,w_t^{1:j},w_t^{j+1})]$, which enjoys a tractable complexity of $O(d\times|a_{t+1}|)$.


\subsection{Per-Token Policy Update}
After defining the soft Bellman backup, we develop an approach for the policy update, i.e. the per-token policy update. According to the definition of ERL~\cite{schulman2017equivalence}, an optimal stochastic policy that refers to $\bar{\pi}$ and the Q-function could be presented as:
\begin{small}
\begin{align}
\pi^*(\cdot|s_t) & = \arg\max_{\pi}\Big\{\mathbb{E}_{a_t\sim\pi}[Q(s_t,a_t)]-\beta D_{KL}[\pi||\bar{\pi}](s_t)\Big\} \nonumber\\
& = \frac{\bar{\pi}(a_t|s_t)\exp(Q(s_t,a_t)/\beta)}{\mathbb{E}_{\bar{a}_t\sim\bar{\pi}(s_t)}[\exp(Q(s_t,\bar{a}_t)/\beta)]}.
\end{align}
Recall the Equation~\ref{action-to-token-within}, we rewrite this optimal policy regarding to the Q function at the token level:
\begin{align}
\pi_Q^*(\cdot|s_t, w_t^{1:j-1})=\frac{\bar{\pi}(w_t^j|s_t,w_t^{1:j-1})\exp(Q(s_t,w_t^{1:j-1},w_t^j)/\beta)}{\mathbb{E}_{\bar{w}_t^j\sim\bar{\pi}}[\exp(Q(s_t,w_t^{1:j-1},\bar{w}_t^j)/\beta)]}.
\end{align}
\end{small}
Due to the discrete nature of vocabularies, we can explicitly calculate the $\pi_Q^*(\cdot|s_t, w_t^{1:j-1})$ with $\bar{\pi}(w_t^j|s_t,w_t^{1:j-1})$ and current Q function $Q(s_t,w_t^{1:j-1},w_t^j)$, 
and thus update the policy toward minimizing the KL-divergence between $\pi(w_t^j|s_t,w_t^{1:j-1})$ and $\pi_Q^*(\cdot|s_t, w_t^{1:j-1})$, which is somewhat similar to the policy update process in SAC~\cite{haarnoja2018soft_applications}. Our per-token policy update scheme is to minimize the objective $\mathcal{J}_{Q}(\pi)$ in Equation~\ref{equ_per_token_policy_update}.

\subsection{A Practical Algorithm}
In practice, we draw a concrete algorithm of ETPO with pseudo-code shown in the Algorithm~\ref{alg_etpo}. Since we don’t want the learned strategy to deviate too far from the original LLM, we constraint the distribution to the original language model $\rho$ by setting the reference policy $\bar{\pi} = \rho$. As for pre-trained LLMs, their output distribution implies their preference for each token, thus the soft Q-network $Q_{\theta}$ is initialized with $\rho$ as well. Besides, Our algorithm makes use of a target soft Q-network $Q_{\bar{\theta}}$ to facilitate a stable learning process, whose parameters are updated softly. 

\begin{algorithm}[tb]
   \caption{ETPO Algorithm.}
   \label{alg_etpo}
\begin{algorithmic}
   \STATE {\bfseries Input:} LLM $\rho$, Learning rate $\alpha$, Polyak coefficient $\lambda$
   \STATE {\bfseries Initialize:} $\pi_{\phi}\gets\rho, \bar{\pi}_{\bar{\phi}}\gets\rho, Q_{\theta}\gets\rho, Q_{\bar{\theta}}\gets\rho$
   \STATE {\bfseries Initialize:} Data buffer $\mathcal{D}\gets\emptyset$
   \FOR{each epoch}
   \FOR{$t=0$ {\bfseries to} $T-1$}
   \STATE Collect $s_t$.
   \STATE $a_t\sim\pi_{\phi}(a_t|s_t)$.
   \STATE $a_{t+1}\sim\mathcal{T}(a_{t+1}|s_t, a_t)$.
   \STATE $\mathcal{D}\gets\mathcal{D}\cup\{(s_t, a_t, r(s_t,a_t),s_{t+1})\}$.
   \ENDFOR
   \STATE Sample a mini-batch $\mathcal{B}$ from $\mathcal{D}$.
   \FOR{each time step $t$ in $\mathcal{B}$}
   \FOR{each token $j$ in $a_t$}
   \STATE $q\gets Q_{\theta}(s_t, w_t^{1:j-1},w_t^j)$
   \IF{$j<|a_t|$}
   \STATE $q_{targ}\gets\mathbb{E}_{w_t^{j+1}\sim\pi_{\phi}(s_t,w_t^{1:j})}[Q_{\bar{\theta}}(s_t,w_t^{1:j},w_t^{j+1})]$
   \STATE \qquad \qquad $-\beta D_{KL}[\pi_{\phi}||\bar{\pi}_{\bar{\phi}}](s_t,w_t^{1:j})$
   \ELSIF{$j=|a_t|$}
   \STATE $q_{targ}\gets r(s_t,a_t)$
   \STATE \qquad \qquad $+\gamma(\mathbb{E}_{w_{t+1}^1\sim\pi_{\phi}(s_{t+1})}[Q_{\bar{\theta}}(s_{t+1},w_{t+1}^1)]$
   \STATE \qquad \qquad $-\beta D_{KL}[\pi_{\phi}||\bar{\pi}_{\bar{\phi}}](s_{t+1})$
   \ENDIF
   \ENDFOR
   \ENDFOR
   \STATE $\theta\gets\theta-\alpha\nabla_{\theta}\mathbb{E}_{\mathcal{B}}[(q-q_{targ})^2]$
   \STATE $\phi\gets\phi-\alpha\nabla_{\phi}\mathbb{E}_{\mathcal{B}}[\mathcal{J}_{Q_{\theta}}(\pi_{\phi})]$
   \STATE $\bar{\theta}\gets\lambda\bar{\theta}+(1-\lambda)\theta$
   \ENDFOR
   \STATE {\bfseries Output:} $\pi_{\phi}$
\end{algorithmic}
\end{algorithm}

\section{Experiments}
\label{sec_experiments}
In this section, we evaluate our method by modeling data science code generation tasks as interactive decision-making environments and training a 7B CodeLlama model with ETPO. Based on the experimental results, we discussed the superiority of our method over several baselines in generating machine learning code to achieve the highest possible ROC AUC score~\cite{narkhede2018understanding} with Scikit-learn module~\cite{pedregosa2011scikit}. 
We also showcase the convergence performance of ETPO, which is important for evaluating RL algorithms. Further, we analyze the difference between the generated code of vanilla CodeLlama-7B and the CodeLlama-7B fine-tuned by ETPO, giving insights about the behavioral emergence of LLMs with RL training. In the end, we examine the perplexity with the benchmarks of Github, Wikitext, and arXiv~\cite{Wu2022MemorizingT} to confirm if ETPO can maintain the model's fundamental capabilities and stability. Ablation studies are conducted in Appendix~\ref{sec_ablation} to investigate the importance of multi-step reflection and per-token soft Bellman update separately.

\subsection{Environmental Setup}
\label{sub_sec_env_setup}
To make the experiment closer to real industrial scenarios, we build an interactive environment that models machine learning code writing and debugging tasks into a sequential decision-making problem. With the same testbeds as \cite{hollmann2023large}, we adopt 4 Kaggle datasets and 10 OpenML datasets~\cite{vanschoren2014openml}, with details in Table~\ref{tab_data_disc}.

\begin{table}[ht]
\vskip -0.15in
\caption{Details of datasets used for experiments, where ⟨\textit{K}⟩ denotes the Kaggle datasets.}
\vskip 0.05in
\begin{center}
\label{tab_data_disc}
\begin{scriptsize}
\begin{sc}
\begin{tabular}{l|c|c|c}
\toprule
Data set & \#Features & \#Samples & \#Classes \\
\midrule
Pharyngitis ⟨\textit{K}⟩  & 19 & 512 & 2 \\
Kidney Stone ⟨\textit{K}⟩ & 7 & 414 & 2\\
Health Insurance ⟨\textit{K}⟩ & 13 & 2000 & 2\\
Spaceship Titanic ⟨\textit{K}⟩ & 13 & 2000 & 2\\
\midrule
Airlines & 7 & 2000 & 2\\
Balance Scale & 4 & 125 & 3\\
Breast-w & 9 & 69 & 2\\
CMC & 9 & 1473 & 3\\
Credit-g & 20 & 1000 & 2\\
Diabetes & 8 & 768 & 2\\
Eucalyptus & 19 & 736 & 5\\
Jungle Chess & 6 & 2000 & 3\\
PC1 & 21 & 1109 & 2\\
Tic-Tac-Toe & 9 & 95 & 2\\
\bottomrule
\end{tabular}
\end{sc}
\end{scriptsize}
\end{center}
\vskip -0.2in
\end{table}

To model the code generation tasks as MDPs, we define $g\in\mathcal{V}^N$ to be the task description, $e\in\mathcal{V}^N$ to be the runtime error message when generated codes are non-runnable, serving as reflection messages.
In the beginning, our agent generates codes as an action $a_0$ based on the initial state $s_0=\eta(g)$ where $\eta$ is a prompting operation, then receives a reward $r$ by executing the codes. The reward function $r(s_t, a_t)$ is defined as:
\begin{align}
\label{equ_reward_function}
r(s, \mathbf{a}_t) =
\begin{cases} 
\mbox{ROC AUC score}, & \mbox{Runnable} \\
-1.0,  & \mbox{Non-runnable} 
\end{cases}
\end{align}
When the codes are executed successfully, a done signal will occur immediately, along with a ROC AUC score $\in[0,1]$. If the codes run wrong, the agent receives a constant $-1.0$ as a reward, as well as the next state $s'=\eta(g, a, e)$. 
Appendix~\ref{appendix_prompts} provides concrete examples of the prompts and generated codes. In addition to successful code execution, a done signal will also be received if $t$ reaches a pre-set maximum step limit, e.g. $5$ in our experiments. 

Adopting the same evaluation metrics as \cite{hollmann2023large}, for each dataset, we evaluate 5 repetitions, each with a different random seed and train-validation split to reduce the variance stemming from these splits~\cite{bouthillier2021accounting}. We split into $50\%$ train and $50\%$ validation samples and all methods used the same splits. We record the highest reward explored by each method within a given number of environment steps $k$ as the main criteria for judgment, similar to evaluating AutoML tasks~\cite{hutter2019automated}.

\subsection{Main Results}

\begin{figure}[!ht]
\begin{center}
\vskip -0.1in
\centerline{\includegraphics[width=0.9\linewidth]{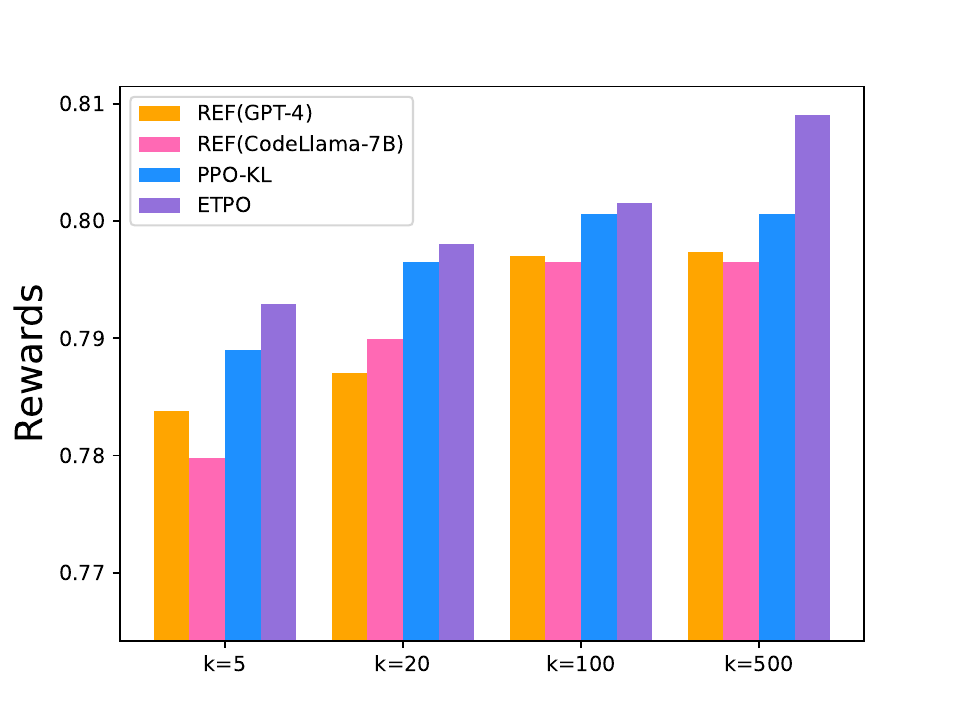}}
\vskip -0.2in
\caption{Average of the best reward explored across 14 datasets corresponding to different environmental steps $k$. "REF" indicates the Reflection baselines.}
\label{fig_average_k}
\end{center}
\vskip -0.3in
\end{figure}

\begin{table*}[!ht]
\vskip -0.2in
\caption{ROC AUC results on validation set where $\pm$ indicates the standard deviation across 5 splits. ⟨\textit{K}⟩ denotes the Kaggle datasets.}
\begin{center}
\label{tab_main_result}
\begin{small}
\begin{sc}
\begin{tabular}{l|cc|cc}
\toprule
\multirow{2}*{Data set} & GPT-4 & CodeLlama-7B & \multicolumn{2}{c}{CodeLlama-7B} \\
\cmidrule{2-5} 
& \multicolumn{2}{c|}{Reflection} & PPO-KL & ETPO \\
\midrule
Pharyngitis ⟨\textit{K}⟩ & 0.6902$\pm$0.008 & 0.7129$\pm$0.010 & 0.7134$\pm$0.009 & \textbf{0.7317}$\pm$0.006 \\
Kidney Stone ⟨\textit{K}⟩ & 0.7833$\pm$0.019 & 0.7825$\pm$0.018 & 0.7840$\pm$0.015 & \textbf{0.7881}$\pm$0.011\\
Health Insurance ⟨\textit{K}⟩ & 0.5826$\pm$0.008 & 0.5725$\pm$0.014 & 0.5849$\pm$0.015 & \textbf{0.5995}$\pm$0.011\\
Spaceship Titanic ⟨\textit{K}⟩ & 0.8532$\pm$0.015 & 0.8524$\pm$0.015 & 0.8575$\pm$0.011 & \textbf{0.8578}$\pm$0.015\\
\midrule
Airlines & 0.6495$\pm$0.013 & 0.6476$\pm$0.012 & 0.6549$\pm$0.011 & \textbf{0.6580}$\pm$0.011\\
Balance Scale & 0.8833$\pm$0.043 & 0.8833$\pm$0.043 & 0.8833$\pm$0.043 & \textbf{0.9289}$\pm$0.038\\
Breast-w & 0.9944$\pm$0.004 & 0.9941$\pm$0.003 & \textbf{0.9948}$\pm$0.003 & \textbf{0.9948}$\pm$0.003\\
CMC & 0.6940$\pm$0.006 & 0.7158$\pm$0.006 & 0.7278$\pm$0.005 & \textbf{0.7306}$\pm$0.003\\
Credit-g & \textbf{0.7883}$\pm$0.024 & 0.7799$\pm$0.021 & 0.7850$\pm$0.025 & 0.7875$\pm$0.023\\
Diabetes & 0.8327$\pm$0.009 & 0.8321$\pm$0.009 & 0.8322$\pm$0.014 & \textbf{0.8329}$\pm$0.008\\
Eucalyptus & 0.8960$\pm$0.006 & 0.8961$\pm$0.004 & 0.8961$\pm$0.004 & \textbf{0.8973}$\pm$0.006\\
Jungle Chess & 0.9100$\pm$0.005 & 0.9097$\pm$0.006 & 0.9107$\pm$0.008 & \textbf{0.9216}$\pm$0.007\\
PC1 & 0.8489$\pm$0.009 & 0.8414$\pm$0.019 & 0.8470$\pm$0.021 & \textbf{0.8492}$\pm$0.016\\
Tic-Tac-Toe & \textbf{0.7560}$\pm$0.039 & 0.7304$\pm$0.057 &  0.7356$\pm$0.036 & 0.7483$\pm$0.032\\
\midrule
Average & 0.7973$\pm$0.015 & 0.7965$\pm$0.017 & 0.8005$\pm$0.016 & \textbf{0.8090}$\pm$0.013\\
\bottomrule
\end{tabular}
\end{sc}
\end{small}
\end{center}
\vskip -0.2in
\end{table*}

Figure~\ref{fig_average_k} shows the trends of average performance of ETPO and three related baseline methods across $14$ datasets under different exploration intensities, i.e. $k$. More specifically, Table~\ref{tab_main_result} shows the highest reward explored by these methods for each dataset within $500$ interactive steps. Among these baselines, the \textit{Reflection} keeps the interactive pipeline of ETPO unchanged but removes the training process as a reference for LLMs' initial performance. 
Besides, similar to RLHF, we also implement an action-level PPO with KL term in the reward function: $r(s_t, a_t)-\beta\log\frac{\pi(a_t|s_t)}{\rho(a_t|s_t)}$, based on the same framework as ETPO to fine-tune CodeLlama-7B model, i.e. \textit{PPO-KL}. As an action-level algorithm, PPO-KL will assign the same credit to all tokens in an action. 

From Figure~\ref{fig_average_k}, we observe that the performance of the Reflections hardly changes as the number of exploration steps increases from $100$ to $500$, which means $k=500$ is enough to touch the caps of the Reflection with vanilla models. Comparing the Reflection with PPO-KL, we confirm that even action-level RL training can slightly improve the model’s ability to generate better code for specific datasets. Further, ETPO outperforms both Reflection and PPO-KL in most of the experimental datasets, which verifies the effectiveness of our method.

\subsection{Convergence}
\label{sub_sec_convergence}

\begin{figure}[!ht]
\vskip -0.1in
\begin{center}
\centerline{\includegraphics[width=0.95\linewidth]{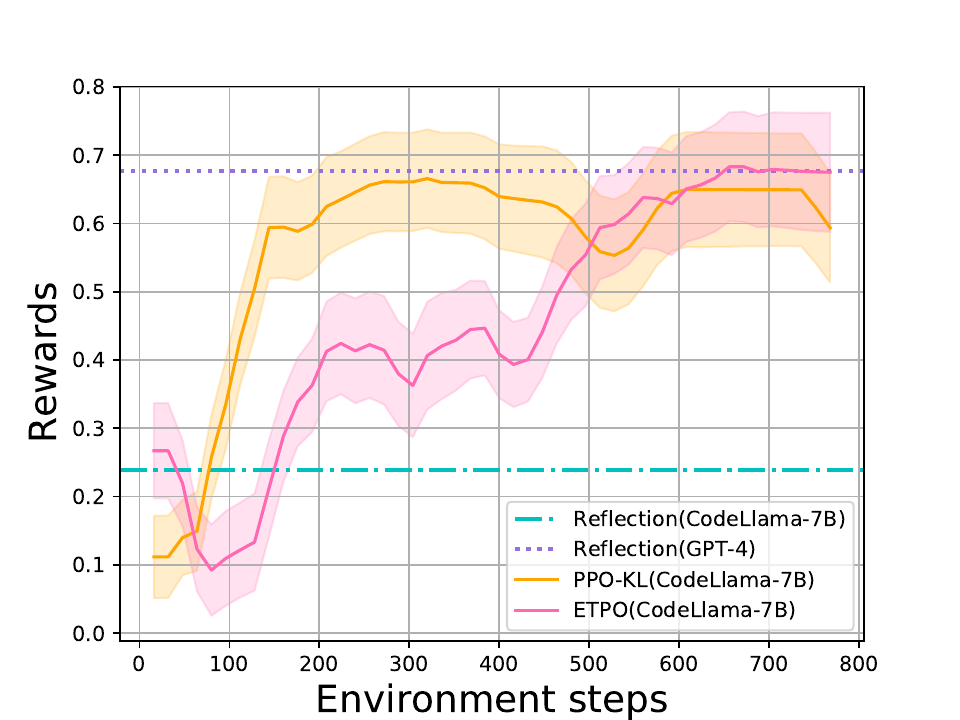}}
\vskip -0.1in
\caption{Average performance comparison during training on all Kaggle datasets that are lesser-known. }
\label{fig_convergence}
\end{center}
\vskip -0.4in
\end{figure}

In this section, we take four Kaggle datasets to further analyze the learning process of PPO-KL and ETPO in depth. To this end, we record the average reward of generated codes in each batch during the training process, instead of the best reward in the previous section. It is also an important indicator for evaluating the convergence performance and stability of RL algorithms~\cite{potapov2003convergence}. Figure~\ref{fig_convergence} shows the average performance on 4 Kaggle datasets achieved by PPO-KL and ETPO as well as the average rewards of Reflection with GPT-4 and CodeLlama-7B after 500 interactions.

According to Figure~\ref{fig_convergence}, although the best codes explored by GPT-4 and CodeLlama-7B enjoy similar rewards for each dataset in the previous section, it does not mean that CodaLlama-7B has equal capabilities to GPT-4. The huge gap between CodeLlama-7B and GPT-4 in average rewards shown in Figure~\ref{fig_convergence} implies that a significant portion of the code generated by CodeLlama-7B is non-runnable and therefore receives a -1 as a reward which brings down the average. This illustrates that vanilla CodeLlama-7B is more unstable than GPT-4 before training. Then, as training progresses, the average performance of PPO-KL and ETPO has improved significantly. Especially PPO-KL, because it allows all tokens in an action to share credit, it can backpropagate the impact of the reward signal to the front end of the token sequence earlier and thus enjoy faster performance improvements. However, the credit assignment granularity of PPO-KL is too large and not precise enough, which prevents further improvement of its performance and brings instability to the convergence process, as shown in the second half of the curve. In contrast, ETPO spreads the impact of reward signals token-by-token. Thus its performance improvement is relatively gentle in the early stage. But, benefit from its fine-grained credit assignment, its improvement process is more stable and converges to average performance exceeding PPO-KL even the Reflection with GPT-4.

\subsection{Behavioural emergence with RL training}
Let’s take a deeper investigation into the behavior of LLM after RL training.
In Figures~\ref{fig_ref_code_demo}~and~\ref{fig_our_code_demo} at Appendix~\ref{sec_gen_code}, we demonstrate the difference of generated codes between the vanilla CodeLlama-7B and the CodeLlama-7B fine-tuned with ETPO. The code generated by the trained model in Figure~\ref{fig_our_code_demo} indicates the effectiveness of entropy-regularized RL in preventing the policy from moving too far and losing the coherence and readability of generated text.

In addition to the difference in code complexity, another significant difference between Figures~\ref{fig_ref_code_demo} and~\ref{fig_our_code_demo} is that the code generated by the vanilla model is more in line with the prompt requirements, i.e. returning a Scikit \textbf{model}, while the code generated by RL trained model deviates from prompt to a certain extent, i.e. it builds a \textbf{pipeline} with data processing operation. This situation is similar to the hallucination~\cite{ji2023survey} issue in the text-generation tasks that needed to be avoided. However, this deviant code in fact achieves higher rewards for the corresponding dataset. From an RL perspective, this situation is easy to understand. Since RL algorithms encourage agents to further explore directions with higher rewards, even unexpected behaviors with high rewards will be continuously enhanced and improved in the subsequent training process. Finally, models are promoted to generate the complex and "noncompliant" code in Figure~\ref{fig_our_code_demo}, which achieves better performance than code generated by repeated exploration without RL training. Figure~\ref{fig_code_iter} gives a more detailed demonstration of how this process occurs.

Further, this phenomenon inspires us an insight that we can filter the hallucination of LLMs with the exploration-exploitation manner~\cite{sutton2018reinforcement} of RL to boost the emergence of new behaviors or capabilities. Limited by token length, it isn't easy to encode all task-related information into prompts. The hallucination could be treated as supplemental explorations in an exploration-exploitation manner. we can filter out those that are beneficial to the specific task and reinforce them to promote new behaviors, instead of deliberately preventing it. On the other way, even those hallucinations that are useless for the current task may still be able to contribute to other tasks with techniques like hindsight relabelling~\cite{li2020generalized} to boost the emergence of new capabilities. In other words, for LLM agents with RL, hallucination is the source of creativity and innovation that drives agents' performance improvement. By embracing the inherent uncertainty and variability in the outputs of LLMs, we can harness their potential with serendipitous discoveries.

\subsection{Affect on Perplexity}
We choose benchmarks Github, Wikitext, and arXiv introduced by memorizing transformers~\cite{Wu2022MemorizingT} to observe the perplexity changes compared to the original model after training by ETPO.

\begin{table}[!ht]
\vskip -0.2in
\caption{Perplexity changes comparison.}
\begin{center}
\label{tab_abl_on_ppl}
\begin{small}
\begin{tabular}{@{}llll@{}}
\toprule
\textbf{} & \textbf{Github} & \textbf{Wikitext} & \textbf{arXiv} \\ \midrule
\textbf{CodeLlama-7B} & 2.0633 & 5.9850 & 3.5768 \\
\textbf{ETPO (Ours)} & 2.0648 & 5.9874 & 3.5787 \\ \bottomrule
\end{tabular}%
\end{small}
\end{center}
\vskip -0.15in
\end{table}

According to the experimental results in Table~\ref{tab_abl_on_ppl}, the minor changes observed after training the CodeLlama-7B model using the ETPO method indicate that this method has a relatively small impact on the original language modeling. This slight variation suggests that although the ETPO method may introduce new optimization or training strategies, it does not fundamentally alter or disrupt the basic logic of language modeling. This is a positive outcome, as it means that while trying to improve model performance or introduce new training methods, the model's fundamental capabilities and stability are maintained.

In the process of training and optimizing language models, maintaining the model's understanding of language structure is crucial. Minor performance improvements or changes indicate that your method may be effective in subtly adjusting the model to suit specific tasks or datasets without sacrificing the model's generality and foundational language abilities. This balance is very important, especially in the development of models that can be broadly applied across different linguistic tasks and domains.

\section{Conclusion}

In this work, we present the ETPO, a token-level optimization approach that leverages entropy-regularized reinforcement learning to train LLM agents for verbal sequential decision-making tasks. ETPO decomposes the optimization in action space to token space and theoretically proves their consistency. Our experiments also confirm the effectiveness of ETPO from different dimensions. Moreover, we acknowledge the existing limitations of ETPO, i.e. the requirement for a quantitative reward function, which is not easily attainable in some textual environment. To mitigate this issue, we envisage integrating ETPO with self-rewarding~\cite{yuan2024self} or hindsight relabeling~\cite{li2020generalized} techniques and leave detailed exploration for future research endeavors.

\clearpage
\section*{Impact Statements}
This paper presents work whose goal is to advance the field of Machine Learning. There are many potential societal consequences of our work, none of which we feel must be specifically highlighted here.


\bibliography{main}

\begin{thebibliography}{52}
\providecommand{\natexlab}[1]{#1}
\providecommand{\url}[1]{\texttt{#1}}
\expandafter\ifx\csname urlstyle\endcsname\relax
  \providecommand{\doi}[1]{doi: #1}\else
  \providecommand{\doi}{doi: \begingroup \urlstyle{rm}\Url}\fi

\bibitem[Bouthillier et~al.(2021)Bouthillier, Delaunay, Bronzi, Trofimov, Nichyporuk, Szeto, Mohammadi~Sepahvand, Raff, Madan, Voleti, et~al.]{bouthillier2021accounting}
Bouthillier, X., Delaunay, P., Bronzi, M., Trofimov, A., Nichyporuk, B., Szeto, J., Mohammadi~Sepahvand, N., Raff, E., Madan, K., Voleti, V., et~al.
\newblock Accounting for variance in machine learning benchmarks.
\newblock \emph{Proceedings of Machine Learning and Systems}, 3:\penalty0 747--769, 2021.

\bibitem[Carta et~al.(2023)Carta, Romac, Wolf, Lamprier, Sigaud, and Oudeyer]{carta2023grounding}
Carta, T., Romac, C., Wolf, T., Lamprier, S., Sigaud, O., and Oudeyer, P.-Y.
\newblock Grounding large language models in interactive environments with online reinforcement learning.
\newblock \emph{arXiv preprint arXiv:2302.02662}, 2023.

\bibitem[Chebotar et~al.(2023)Chebotar, Vuong, Hausman, Xia, Lu, Irpan, Kumar, Yu, Herzog, Pertsch, et~al.]{chebotar2023q}
Chebotar, Y., Vuong, Q., Hausman, K., Xia, F., Lu, Y., Irpan, A., Kumar, A., Yu, T., Herzog, A., Pertsch, K., et~al.
\newblock Q-transformer: Scalable offline reinforcement learning via autoregressive q-functions.
\newblock In \emph{Conference on Robot Learning}, pp.\  3909--3928. PMLR, 2023.

\bibitem[Chen et~al.(2024)Chen, Zhou, Zhao, Wan, Zhang, Zhang, and Wen]{chen2024improving}
Chen, Z., Zhou, K., Zhao, W.~X., Wan, J., Zhang, F., Zhang, D., and Wen, J.-R.
\newblock Improving large language models via fine-grained reinforcement learning with minimum editing constraint.
\newblock \emph{arXiv preprint arXiv:2401.06081}, 2024.

\bibitem[Deng et~al.(2023)Deng, Zhang, He, Chen, Shi, Zhou, Fu, Zhang, Wang, Zhou, Lin, and He]{Deng2023K2AF}
Deng, C., Zhang, T., He, Z., Chen, Q., Shi, Y., Zhou, L., Fu, L., Zhang, W., Wang, X., Zhou, C., Lin, Z., and He, J.
\newblock K2: A foundation language model for geoscience knowledge understanding and utilization.
\newblock 2023.
\newblock URL \url{https://api.semanticscholar.org/CorpusID:259108887}.

\bibitem[Feng et~al.(2023)Feng, Wan, Wen, Wen, Zhang, and Wang]{feng2023alphazero}
Feng, X., Wan, Z., Wen, M., Wen, Y., Zhang, W., and Wang, J.
\newblock Alphazero-like tree-search can guide large language model decoding and training.
\newblock \emph{arXiv preprint arXiv:2309.17179}, 2023.

\bibitem[Fox et~al.(2015)Fox, Pakman, and Tishby]{fox2015taming}
Fox, R., Pakman, A., and Tishby, N.
\newblock Taming the noise in reinforcement learning via soft updates.
\newblock \emph{arXiv preprint arXiv:1512.08562}, 2015.

\bibitem[Haarnoja et~al.(2017)Haarnoja, Tang, Abbeel, and Levine]{haarnoja2017reinforcement}
Haarnoja, T., Tang, H., Abbeel, P., and Levine, S.
\newblock Reinforcement learning with deep energy-based policies.
\newblock In \emph{International conference on machine learning}, pp.\  1352--1361. PMLR, 2017.

\bibitem[Haarnoja et~al.(2018{\natexlab{a}})Haarnoja, Zhou, Abbeel, and Levine]{haarnoja2018soft}
Haarnoja, T., Zhou, A., Abbeel, P., and Levine, S.
\newblock Soft actor-critic: Off-policy maximum entropy deep reinforcement learning with a stochastic actor.
\newblock In \emph{International conference on machine learning}, pp.\  1861--1870. PMLR, 2018{\natexlab{a}}.

\bibitem[Haarnoja et~al.(2018{\natexlab{b}})Haarnoja, Zhou, Hartikainen, Tucker, Ha, Tan, Kumar, Zhu, Gupta, Abbeel, et~al.]{haarnoja2018soft_applications}
Haarnoja, T., Zhou, A., Hartikainen, K., Tucker, G., Ha, S., Tan, J., Kumar, V., Zhu, H., Gupta, A., Abbeel, P., et~al.
\newblock Soft actor-critic algorithms and applications.
\newblock \emph{arXiv preprint arXiv:1812.05905}, 2018{\natexlab{b}}.

\bibitem[Hao et~al.(2023)Hao, Gu, Ma, Hong, Wang, Wang, and Hu]{hao2023reasoning}
Hao, S., Gu, Y., Ma, H., Hong, J.~J., Wang, Z., Wang, D.~Z., and Hu, Z.
\newblock Reasoning with language model is planning with world model.
\newblock \emph{arXiv preprint arXiv:2305.14992}, 2023.

\bibitem[Hollmann et~al.(2023)Hollmann, M{\"u}ller, and Hutter]{hollmann2023large}
Hollmann, N., M{\"u}ller, S., and Hutter, F.
\newblock Large language models for automated data science: Introducing caafe for context-aware automated feature engineering.
\newblock In \emph{Thirty-seventh Conference on Neural Information Processing Systems}, 2023.

\bibitem[Hutter et~al.(2019)Hutter, Kotthoff, and Vanschoren]{hutter2019automated}
Hutter, F., Kotthoff, L., and Vanschoren, J.
\newblock \emph{Automated machine learning: methods, systems, challenges}.
\newblock Springer Nature, 2019.

\bibitem[Ji et~al.(2023)Ji, Lee, Frieske, Yu, Su, Xu, Ishii, Bang, Madotto, and Fung]{ji2023survey}
Ji, Z., Lee, N., Frieske, R., Yu, T., Su, D., Xu, Y., Ishii, E., Bang, Y.~J., Madotto, A., and Fung, P.
\newblock Survey of hallucination in natural language generation.
\newblock \emph{ACM Computing Surveys}, 55\penalty0 (12):\penalty0 1--38, 2023.

\bibitem[Kappen(2011)]{kappen2011optimal}
Kappen, H.~J.
\newblock Optimal control theory and the linear bellman equation.
\newblock \emph{Cambridge: Cambridge University Press}, 2011.

\bibitem[Kappen et~al.(2012)Kappen, G{\'o}mez, and Opper]{kappen2012optimal}
Kappen, H.~J., G{\'o}mez, V., and Opper, M.
\newblock Optimal control as a graphical model inference problem.
\newblock \emph{Machine learning}, 87:\penalty0 159--182, 2012.

\bibitem[Koller \& Friedman(2009)Koller and Friedman]{koller2009probabilistic}
Koller, D. and Friedman, N.
\newblock \emph{Probabilistic graphical models: principles and techniques}.
\newblock MIT press, 2009.

\bibitem[Konda \& Tsitsiklis(1999)Konda and Tsitsiklis]{konda1999actor}
Konda, V. and Tsitsiklis, J.
\newblock Actor-critic algorithms.
\newblock \emph{Advances in neural information processing systems}, 12, 1999.

\bibitem[Le et~al.(2022)Le, Wang, Gotmare, Savarese, and Hoi]{le2022coderl}
Le, H., Wang, Y., Gotmare, A.~D., Savarese, S., and Hoi, S. C.~H.
\newblock Coderl: Mastering code generation through pretrained models and deep reinforcement learning.
\newblock \emph{Advances in Neural Information Processing Systems}, 35:\penalty0 21314--21328, 2022.

\bibitem[Levine(2018)]{levine2018reinforcement}
Levine, S.
\newblock Reinforcement learning and control as probabilistic inference: Tutorial and review.
\newblock \emph{arXiv preprint arXiv:1805.00909}, 2018.

\bibitem[Li et~al.(2020)Li, Pinto, and Abbeel]{li2020generalized}
Li, A., Pinto, L., and Abbeel, P.
\newblock Generalized hindsight for reinforcement learning.
\newblock \emph{Advances in neural information processing systems}, 33:\penalty0 7754--7767, 2020.

\bibitem[Lin et~al.(2023)Lin, Deng, Zhou, Zhang, Xu, Xu, He, Shi, Dai, Song, Zeng, Chen, Shi, Huang, Xu, Wang, Fu, Zhang, He, Ma, Zhu, Wang, and Zhou]{Lin2023GeoGalacticaAS}
Lin, Z., Deng, C., Zhou, L., Zhang, T., Xu, Y., Xu, Y., He, Z., Shi, Y., Dai, B., Song, Y., Zeng, B., Chen, Q., Shi, T., Huang, T., Xu, Y., Wang, S., Fu, L., Zhang, W., He, J., Ma, C., Zhu, Y., Wang, X., and Zhou, C.
\newblock Geogalactica: A scientific large language model in geoscience.
\newblock \emph{ArXiv}, abs/2401.00434, 2023.
\newblock URL \url{https://api.semanticscholar.org/CorpusID:266693296}.

\bibitem[Madaan et~al.(2023)Madaan, Tandon, Gupta, Hallinan, Gao, Wiegreffe, Alon, Dziri, Prabhumoye, Yang, et~al.]{madaan2023self}
Madaan, A., Tandon, N., Gupta, P., Hallinan, S., Gao, L., Wiegreffe, S., Alon, U., Dziri, N., Prabhumoye, S., Yang, Y., et~al.
\newblock Self-refine: Iterative refinement with self-feedback.
\newblock \emph{arXiv preprint arXiv:2303.17651}, 2023.

\bibitem[Nachum et~al.(2017)Nachum, Norouzi, Xu, and Schuurmans]{nachum2017bridging}
Nachum, O., Norouzi, M., Xu, K., and Schuurmans, D.
\newblock Bridging the gap between value and policy based reinforcement learning.
\newblock \emph{Advances in neural information processing systems}, 30, 2017.

\bibitem[Narkhede(2018)]{narkhede2018understanding}
Narkhede, S.
\newblock Understanding auc-roc curve.
\newblock \emph{Towards data science}, 26\penalty0 (1):\penalty0 220--227, 2018.

\bibitem[Ouyang et~al.(2022)Ouyang, Wu, Jiang, Almeida, Wainwright, Mishkin, Zhang, Agarwal, Slama, Ray, et~al.]{ouyang2022training}
Ouyang, L., Wu, J., Jiang, X., Almeida, D., Wainwright, C., Mishkin, P., Zhang, C., Agarwal, S., Slama, K., Ray, A., et~al.
\newblock Training language models to follow instructions with human feedback.
\newblock \emph{Advances in Neural Information Processing Systems}, 35:\penalty0 27730--27744, 2022.

\bibitem[Pedregosa et~al.(2011)Pedregosa, Varoquaux, Gramfort, Michel, Thirion, Grisel, Blondel, Prettenhofer, Weiss, Dubourg, et~al.]{pedregosa2011scikit}
Pedregosa, F., Varoquaux, G., Gramfort, A., Michel, V., Thirion, B., Grisel, O., Blondel, M., Prettenhofer, P., Weiss, R., Dubourg, V., et~al.
\newblock Scikit-learn: Machine learning in python.
\newblock \emph{the Journal of machine Learning research}, 12:\penalty0 2825--2830, 2011.

\bibitem[Potapov \& Ali(2003)Potapov and Ali]{potapov2003convergence}
Potapov, A. and Ali, M.
\newblock Convergence of reinforcement learning algorithms and acceleration of learning.
\newblock \emph{Physical Review E}, 67\penalty0 (2):\penalty0 026706, 2003.

\bibitem[Pryzant et~al.(2023)Pryzant, Iter, Li, Lee, Zhu, and Zeng]{pryzant2023automatic}
Pryzant, R., Iter, D., Li, J., Lee, Y.~T., Zhu, C., and Zeng, M.
\newblock Automatic prompt optimization with" gradient descent" and beam search.
\newblock \emph{arXiv preprint arXiv:2305.03495}, 2023.

\bibitem[Qian et~al.(2023)Qian, Cong, Yang, Chen, Su, Xu, Liu, and Sun]{qian2023communicative}
Qian, C., Cong, X., Yang, C., Chen, W., Su, Y., Xu, J., Liu, Z., and Sun, M.
\newblock Communicative agents for software development.
\newblock \emph{arXiv preprint arXiv:2307.07924}, 2023.

\bibitem[Radford et~al.(2018)Radford, Narasimhan, Salimans, Sutskever, et~al.]{radford2018improving}
Radford, A., Narasimhan, K., Salimans, T., Sutskever, I., et~al.
\newblock Improving language understanding by generative pre-training.
\newblock 2018.

\bibitem[Ramamurthy et~al.(2022)Ramamurthy, Ammanabrolu, Brantley, Hessel, Sifa, Bauckhage, Hajishirzi, and Choi]{ramamurthy2022reinforcement}
Ramamurthy, R., Ammanabrolu, P., Brantley, K., Hessel, J., Sifa, R., Bauckhage, C., Hajishirzi, H., and Choi, Y.
\newblock Is reinforcement learning (not) for natural language processing?: Benchmarks, baselines, and building blocks for natural language policy optimization.
\newblock \emph{arXiv preprint arXiv:2210.01241}, 2022.

\bibitem[Romera-Paredes et~al.(2023)Romera-Paredes, Barekatain, Novikov, Balog, Kumar, Dupont, Ruiz, Ellenberg, Wang, Fawzi, et~al.]{romera2023mathematical}
Romera-Paredes, B., Barekatain, M., Novikov, A., Balog, M., Kumar, M.~P., Dupont, E., Ruiz, F.~J., Ellenberg, J.~S., Wang, P., Fawzi, O., et~al.
\newblock Mathematical discoveries from program search with large language models.
\newblock \emph{Nature}, pp.\  1--3, 2023.

\bibitem[Schick et~al.(2023)Schick, Dwivedi-Yu, Dess{\`\i}, Raileanu, Lomeli, Zettlemoyer, Cancedda, and Scialom]{schick2023toolformer}
Schick, T., Dwivedi-Yu, J., Dess{\`\i}, R., Raileanu, R., Lomeli, M., Zettlemoyer, L., Cancedda, N., and Scialom, T.
\newblock Toolformer: Language models can teach themselves to use tools.
\newblock \emph{arXiv preprint arXiv:2302.04761}, 2023.

\bibitem[Schulman et~al.(2017{\natexlab{a}})Schulman, Chen, and Abbeel]{schulman2017equivalence}
Schulman, J., Chen, X., and Abbeel, P.
\newblock Equivalence between policy gradients and soft q-learning.
\newblock \emph{arXiv preprint arXiv:1704.06440}, 2017{\natexlab{a}}.

\bibitem[Schulman et~al.(2017{\natexlab{b}})Schulman, Wolski, Dhariwal, Radford, and Klimov]{schulman2017proximal}
Schulman, J., Wolski, F., Dhariwal, P., Radford, A., and Klimov, O.
\newblock Proximal policy optimization algorithms.
\newblock \emph{arXiv preprint arXiv:1707.06347}, 2017{\natexlab{b}}.

\bibitem[Shinn et~al.(2023)Shinn, Labash, and Gopinath]{shinn2023reflexion}
Shinn, N., Labash, B., and Gopinath, A.
\newblock Reflexion: an autonomous agent with dynamic memory and self-reflection.
\newblock \emph{arXiv preprint arXiv:2303.11366}, 2023.

\bibitem[Shridhar et~al.(2020)Shridhar, Yuan, C{\^o}t{\'e}, Bisk, Trischler, and Hausknecht]{shridhar2020alfworld}
Shridhar, M., Yuan, X., C{\^o}t{\'e}, M.-A., Bisk, Y., Trischler, A., and Hausknecht, M.
\newblock Alfworld: Aligning text and embodied environments for interactive learning.
\newblock \emph{arXiv preprint arXiv:2010.03768}, 2020.

\bibitem[Sutton \& Barto(2018)Sutton and Barto]{sutton2018reinforcement}
Sutton, R.~S. and Barto, A.~G.
\newblock \emph{Reinforcement learning: An introduction}.
\newblock MIT press, 2018.

\bibitem[Todorov(2008)]{todorov2008general}
Todorov, E.
\newblock General duality between optimal control and estimation.
\newblock In \emph{2008 47th IEEE Conference on Decision and Control}, pp.\  4286--4292. IEEE, 2008.

\bibitem[Toussaint(2009)]{toussaint2009robot}
Toussaint, M.
\newblock Robot trajectory optimization using approximate inference.
\newblock In \emph{Proceedings of the 26th annual international conference on machine learning}, pp.\  1049--1056, 2009.

\bibitem[Touvron et~al.(2023)Touvron, Martin, Stone, Albert, Almahairi, Babaei, Bashlykov, Batra, Bhargava, Bhosale, et~al.]{touvron2023llama}
Touvron, H., Martin, L., Stone, K., Albert, P., Almahairi, A., Babaei, Y., Bashlykov, N., Batra, S., Bhargava, P., Bhosale, S., et~al.
\newblock Llama 2: Open foundation and fine-tuned chat models.
\newblock \emph{arXiv preprint arXiv:2307.09288}, 2023.

\bibitem[Van~Otterlo \& Wiering(2012)Van~Otterlo and Wiering]{van2012reinforcement}
Van~Otterlo, M. and Wiering, M.
\newblock Reinforcement learning and markov decision processes.
\newblock In \emph{Reinforcement learning: State-of-the-art}, pp.\  3--42. Springer, 2012.

\bibitem[Vanschoren et~al.(2014)Vanschoren, Van~Rijn, Bischl, and Torgo]{vanschoren2014openml}
Vanschoren, J., Van~Rijn, J.~N., Bischl, B., and Torgo, L.
\newblock Openml: networked science in machine learning.
\newblock \emph{ACM SIGKDD Explorations Newsletter}, 15\penalty0 (2):\penalty0 49--60, 2014.

\bibitem[Wen et~al.(2022)Wen, Kuba, Lin, Zhang, Wen, Wang, and Yang]{wen2022multi}
Wen, M., Kuba, J., Lin, R., Zhang, W., Wen, Y., Wang, J., and Yang, Y.
\newblock Multi-agent reinforcement learning is a sequence modeling problem.
\newblock \emph{Advances in Neural Information Processing Systems}, 35:\penalty0 16509--16521, 2022.

\bibitem[Wen et~al.(2024)Wen, Wan, Zhang, Wang, and Wen]{wen2024reinforcing}
Wen, M., Wan, Z., Zhang, W., Wang, J., and Wen, Y.
\newblock Reinforcing language agents via policy optimization with action decomposition, 2024.

\bibitem[Wu et~al.(2022)Wu, Rabe, Hutchins, and Szegedy]{Wu2022MemorizingT}
Wu, Y., Rabe, M.~N., Hutchins, D.~S., and Szegedy, C.
\newblock Memorizing transformers.
\newblock \emph{ArXiv}, abs/2203.08913, 2022.
\newblock URL \url{https://api.semanticscholar.org/CorpusID:247519194}.

\bibitem[Yao et~al.(2022)Yao, Zhao, Yu, Du, Shafran, Narasimhan, and Cao]{yao2022react}
Yao, S., Zhao, J., Yu, D., Du, N., Shafran, I., Narasimhan, K., and Cao, Y.
\newblock React: Synergizing reasoning and acting in language models.
\newblock \emph{arXiv preprint arXiv:2210.03629}, 2022.

\bibitem[Yuan et~al.(2024)Yuan, Pang, Cho, Sukhbaatar, Xu, and Weston]{yuan2024self}
Yuan, W., Pang, R.~Y., Cho, K., Sukhbaatar, S., Xu, J., and Weston, J.
\newblock Self-rewarding language models.
\newblock \emph{arXiv preprint arXiv:2401.10020}, 2024.

\bibitem[Zhang et~al.(2018)Zhang, Vinyals, Munos, and Bengio]{zhang2018study}
Zhang, C., Vinyals, O., Munos, R., and Bengio, S.
\newblock A study on overfitting in deep reinforcement learning.
\newblock \emph{arXiv preprint arXiv:1804.06893}, 2018.

\bibitem[Ziebart(2010)]{ziebart2010modeling}
Ziebart, B.~D.
\newblock \emph{Modeling purposeful adaptive behavior with the principle of maximum causal entropy}.
\newblock Carnegie Mellon University, 2010.

\bibitem[Ziegler et~al.(2019)Ziegler, Stiennon, Wu, Brown, Radford, Amodei, Christiano, and Irving]{ziegler2019fine}
Ziegler, D.~M., Stiennon, N., Wu, J., Brown, T.~B., Radford, A., Amodei, D., Christiano, P., and Irving, G.
\newblock Fine-tuning language models from human preferences.
\newblock \emph{arXiv preprint arXiv:1909.08593}, 2019.

\end{thebibliography}
\bibliographystyle{icml2024}

\newpage
\appendix
\onecolumn
\section{Proof of Optimization Consistency}
\label{appendix_soft_q_update}
We show that optimizing the entropy-argument soft Q-function with our per-token soft Bellman update is consistent with optimizing for the full action. We denote $a = (w_1, w_2,\dots,w_{|a|})$ and $s', a'$ as the next state and action for convenience. We also adopt $\text{KL}(a|s)=D_{KL}[\pi^*||\bar{\pi}](s)$, where the $\text{KL}(a|s)$ indicate it is a action level KL divergence, $\text{KL}(w|s)$ indicate a token level KL divergence.

Given the optimal stochastic policy, $\pi^*$, the vanilla soft Bellman backup for full actions is:
\begin{align}
\label{action-level-update}
\mathbb{E}_{a\sim\pi^*(s)}[Q(s,a)] & = \mathbb{E}_{a\sim\pi^*(s)}\Big[r(s,a)+\gamma(\mathbb{E}_{a'\sim\pi^*(s')}[Q(s',a')]-\beta\text{KL}(a'|s'))\Big] \nonumber\\
& = \mathbb{E}_{a\sim\pi^*(s)}[r(s,a)]+\gamma\mathbb{E}_{a\sim\pi^*(s),s'\sim\mathcal{T}(s,a),a'\sim\pi^*(s')}[Q(s',a')]- \gamma\beta\mathbb{E}_{a\sim\pi^*(s),s'\sim\mathcal{T}(s,a)}[\text{KL}(a'|s')].
\end{align}

Following our Per-token Soft Bellman update, the update over each token within an action ($j<|a|$) is:
\begin{align}
\mathbb{E}_{w_1\sim\pi^*(s)}[Q(s,w_1)] & = \mathbb{E}_{w_1\sim\pi^*(s)}[\mathbb{E}_{w_2\sim\pi^*(s,w_1)}[Q(s,w_1,w_2)]-\beta\text{KL}(w_2|s,w_1)] \nonumber\\
& = \mathbb{E}_{w_1\sim\pi^*(s),w_2\sim\pi^*(s,w_1)}[Q(s,w_1,w_2)]-\beta\mathbb{E}_{w_1\sim\pi^*(s)}[\text{KL}(w_2|s,w_1)] \nonumber\\
& = \mathbb{E}_{w_1\sim\pi^*(s),\dots, w_{|a|}\sim\pi^*(s,w_{1:|a|-1})}[Q(s,w_{1:|a|-1},w_{|a|})] \nonumber\\
&~~~~ - \beta\mathbb{E}_{w_1\sim\pi^*(s),\dots, w_{|a|-1}\sim\pi^*(s,w_{1:|a|-2})}[\sum_{j=1}^{|a|-1}\text{KL}(w_{j+1}|s,w_{1:j})].
\end{align}

Minus $\beta\text{KL}(w_1|s)$ from both sides (Note here we use $\mathbb{E}_{a\sim\pi^*(s)}[\cdot]=\mathbb{E}_{w_1\sim\pi^*(s),\dots, w_{|a|}\sim\pi^*(s,w_{1:|a|-1})}[\cdot]$ for convenience), we have:
\begin{align}
\label{action-to-token-within}
\mathbb{E}_{w_1\sim\pi^*(s)}[Q(s,w_1)]-\beta\text{KL}(w_1|s) = \mathbb{E}_{a\sim\pi^*(s)}[Q(s,a)]-\beta\text{KL}(a|s),
\end{align}
where $a=w_{1:|a|}$ and $\text{KL}(a|s)=\sum_{j=1}^{|a|}\text{KL}(w_j|s, w_{1:j-1})$.

Then, the update across the current action $a$ and the next action $a'$ with our Per-token Soft Bellman update is:
\begin{align}
\label{token-level-update}
\mathbb{E}_{a\sim\pi^*(s)}[Q(s,a)] & = \mathbb{E}_{a\sim\pi^*(s)}[r(s,a)] + \gamma(\mathbb{E}_{a\sim\pi^*(s),s'\sim\mathcal{T}(s,a),w'_1\sim\pi^*(s')}[Q(s',w'_1)]-\beta\mathbb{E}_{a\sim\pi^*(s),s'\sim\mathcal{T}(s,a)}[\text{KL}(w'_1|s')])\nonumber\\
& = \mathbb{E}_{a\sim\pi^*(s)}[r(s,a)]+\gamma\mathbb{E}_{a\sim\pi^*(s),s'\sim\mathcal{T}(s,a),a'\sim\pi^*(s')}[Q(s',a')]-\gamma\beta\mathbb{E}_{a\sim\pi^*(s),s'\sim\mathcal{T}(s,a)}[\text{KL}(a'|s')].
\end{align}

Eq.\ref{token-level-update} enjoys the same shape as Eq.\ref{action-level-update}, thus optimizing the soft Q function following our per-token soft Bellman update is equivalent to the action-level optimization. We prove the consistency between per-token soft Bellman updates and soft Bellman updates over full actions.
\section{Per-token Policy Update}
\label{appendix_policy_update}
Optimal Soft Policy w.r.t Q function:
\begin{align}
\pi^*(a|s) & = \arg\max_{\pi}\Big\{\mathbb{E}_{a\sim\pi(s)}[Q(s,a)]-\beta\text{KL}(s)\Big\} \nonumber\\
& = \rho(a|s)\exp(Q(s,a)/\beta)/\mathbb{E}_{a'\sim\rho(s)}[\exp(Q(s,a')/\beta)].\nonumber
\end{align}

Recall Eq.\ref{action-to-token-within}, the optimal soft policy can also be extended to token level:
\begin{align}
\pi^*(w_j|s,w_{1:j-1})=\rho(w_j|s, w_{1:j-1})\exp(Q(s,w_{1:j-1}, w_j)/\beta)/\mathbb{E}_{\bar{w}_j\sim\rho(s, w_{1:j-1})}[\exp(Q(s,w_{1:j-1}, \bar{w}_j)/\beta)]\nonumber
\end{align}

Since $\pi^*(w_j|s,w_{1:j-1})\propto\exp(Q(s,w_{1:j-1}, w_j)/\beta)$, we can update the policy toward minimizing the KL-Divergence between $\pi(w_j|s,w_{1:j-1})$ and $\exp(Q(s,w_{1:j-1}, w_j)/\beta)$:
\begin{align}
\pi_{\text{new}}(w_j|s,w_{1:j-1})=\arg\min_{\pi}\text{KL}\Big[\pi(w_j|s,w_{1:j-1})||\exp(Q(s,w_{1:j-1}, w_j)/\beta)\Big].
\end{align}

In practice, we only take one gradient step on this objective, similar to the policy improvement step in SAC.

\clearpage

\section{Details Pipeline of Experiment}

\begin{figure*}[!ht]
\begin{center}
\centerline{\includegraphics[width=\linewidth]{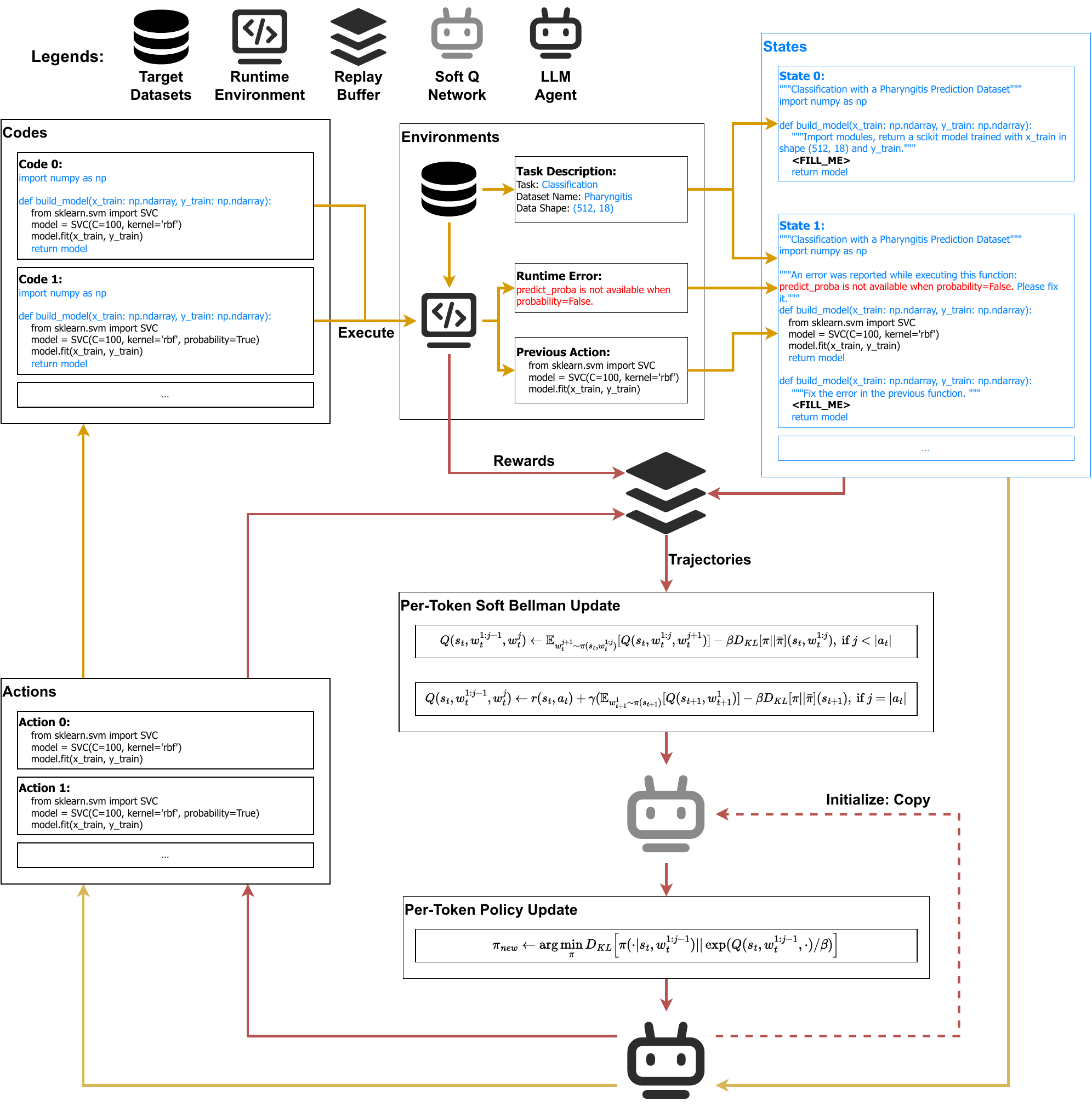}}
\vskip -0.1in
\caption{Detailed pipeline of applying ETPO to the data science code generation task, where the yellow curve demonstrates the sampling process and the red curve indicates the training loop. The dotted line means the soft Q network is initialized with the policy network, i.e. the LLM, at the beginning. The code generated by LLM agents will be used to replace the ``\textbf{FILL\_ME}'' component in the states.}
\label{fig_main_old} 
\vskip -0.3in
\end{center}
\end{figure*}

\clearpage

\section{Simple Prompts}
\label{appendix_prompts}
\begin{figure}[!ht]
    \centering
    \begin{minipage}{\linewidth}
        \centering
\begin{lstlisting}[language=diff,linewidth=  {\linewidth},frame=tb,basicstyle=\footnotesize\ttfamily]
"""Classification with a Pharyngitis Prediction Dataset"""
import numpy as np

def build_model(x_train: np.ndarray, y_train: np.ndarray):
    """Import modules, return a scikit model trained with x_train in shape (256, 18)."""
    from sklearn.svm import SVC
    model = SVC(C=100, kernel='rbf')
    model.fit(x_train, y_train)
    return model
\end{lstlisting}
        \vspace{-10pt}
        \caption{A demonstration of prompts and generated code at the initial time step in our experiment. The text in {\color{blue}blue} is the prompted task description $\eta(g)$ input to LLMs and the code they output is shown in black.}
        \label{fig_first_prompt}
    \vspace{15pt}
    \end{minipage}
    \begin{minipage}{\linewidth}
        \centering
\begin{lstlisting}[language=diff,linewidth=  {\linewidth},frame=tb,basicstyle=\footnotesize\ttfamily]
"""Classification with a Pharyngitis Prediction Dataset"""
import numpy as np

"""An error was reported while executing this function: predict_proba is not available when probability=False. Please fix it."""
# def build_model(x_train: np.ndarray, y_train: np.ndarray):
#    """Import modules, return a scikit model trained with x_train in shape (256, 18)."""
#    from sklearn.svm import SVC
#    model = SVC(C=100, kernel='rbf')
#    model.fit(x_train, y_train)
#    return model

def build_model(x_train: np.ndarray, y_train: np.ndarray):
    """Fix the error in the previous function. """
    from sklearn.svm import SVC
    model = SVC(C=100, kernel='rbf', probability=True)
    model.fit(x_train, y_train)
    return model
\end{lstlisting}
        \vspace{-10pt}
        \caption{A demonstration of prompts and generated code at reflective steps in our experiment. The text in color is the prompt input to LLMs and the code they output is shown in black. More specifically, a runtime error message $e_t$ is decorated with {\color{red}red} while a previous action $a_{t-1}$ is decorated with {\color{codegreen}green}.}
        \label{fig_reflection_prompt}
    \end{minipage}
\end{figure}

\clearpage

\section{Generated Code Examples}
\label{sec_gen_code}
\begin{figure}[!ht]
    \centering
    \begin{minipage}{\linewidth}
        \centering
\begin{lstlisting}[language=diff,linewidth=  {\linewidth},frame=tb,basicstyle=\footnotesize\ttfamily]
"""Classification with a Balance Scale Weight & Distance Prediction Dataset"""
import numpy as np

def build_model(x_train: np.ndarray, y_train: np.ndarray):
    """Import modules, return a scikit model trained with x_train in shape (62, 4) and y_train."""
    
    from sklearn.linear_model import LogisticRegression
    model = LogisticRegression()
    model.fit(x_train, y_train)
    return model
\end{lstlisting}
        \vspace{-10pt}
        \caption{A demonstration of the best code generated by vanilla CodeLlama-7B for the Balance Scale dataset in 500 exploration steps. It achieves the ROC AUC score of 0.8833.}
        \label{fig_ref_code_demo}
    \vspace{15pt}
    \end{minipage}
    \begin{minipage}{\linewidth}
        \centering
\begin{lstlisting}[language=diff,linewidth=  {\linewidth},frame=tb,basicstyle=\footnotesize\ttfamily]
"""Classification with a Balance Scale Weight & Distance Prediction Dataset"""
import numpy as np

def build_model(x_train: np.ndarray, y_train: np.ndarray):
    """Import modules, return a scikit model trained with x_train in shape (62, 4) and y_train."""
    from sklearn.pipeline import Pipeline
    from sklearn.preprocessing import StandardScaler
    from sklearn.neural_network import MLPClassifier
    
    pipeline = Pipeline([
        ('scaler', StandardScaler()),
        ('model', MLPClassifier(hidden_layer_sizes=(100,), activation='relu', solver='lbfgs', max_iter=1000, alpha=1e-5,
            batch_size=100, learning_rate_init=0.1, learning_rate='adaptive', tol=1e-4, verbose=True,
            warm_start=False, momentum=0.9, nesterovs_momentum=True, early_stopping=True,
            validation_fraction=0.1, beta_1=0.9, beta_2=0.999, epsilon=1e-8, n_iter_no_change=10,
            max_fun=15000)),
    ])
    
    model = pipeline.fit(x_train, y_train)
    
    return model
\end{lstlisting}
        \vspace{-10pt}
        \caption{A demonstration of the best code generated by CodeLlama-7B after RL training for the Balance Scale dataset. It achieves the ROC AUC score of 0.9289.}
        \label{fig_our_code_demo}
    \end{minipage}
\end{figure}

\clearpage

\section{Additional Results}

\begin{table*}[!ht]
\label{tab_ablations}
\vskip -0.1in
\caption{Details of performance comparison between different ablations across 14 datasets.}
\vskip 0.05in
\begin{center}
\begin{sc}
\begin{tabular}{l|c|c|c}
\toprule
Data set & ETPO(Disc.) & ETPO(1-Step) & ETPO \\
\midrule
Pharyngitis ⟨\textit{K}⟩ & 0.7134$\pm$0.009 & \textbf{0.7357}$\pm$0.007 & 0.7317$\pm$0.006\\
Kidney Stone ⟨\textit{K}⟩ & \textbf{0.7881}$\pm$0.011 & 0.7861$\pm$0.012 & \textbf{0.7881}$\pm$0.011\\
Health Insurance ⟨\textit{K}⟩ & 0.5849$\pm$0.015 & 0.5829$\pm$0.016 & \textbf{0.5995}$\pm$0.011\\
Spaceship Titanic ⟨\textit{K}⟩ & 0.8574$\pm$0.013 & \textbf{0.8597}$\pm$0.013 & 0.8578$\pm$0.015\\
\midrule
Airlines & 0.6540$\pm$0.011 & 0.6564$\pm$0.010 & \textbf{0.6580}$\pm$0.011\\
Balance Scale & 0.8966$\pm$0.066 & 0.9020$\pm$0.043 & \textbf{0.9289}$\pm$0.038\\
Breast-w & \textbf{0.9948}$\pm$0.003 & 0.9941$\pm$0.004 & \textbf{0.9948}$\pm$0.003\\
CMC & 0.7278$\pm$0.005 & 0.7296$\pm$0.004 & \textbf{0.7306}$\pm$0.003\\
Credit-g & 0.7850$\pm$0.025 & 0.7868$\pm$0.024 & \textbf{0.7875}$\pm$0.023\\
Diabetes & 0.8327$\pm$0.009 & 0.8327$\pm$0.009 & \textbf{0.8329}$\pm$0.008\\
Eucalyptus & 0.8961$\pm$0.004 & 0.8961$\pm$0.004 & \textbf{0.8973}$\pm$0.006\\
Jungle Chess & 0.9100$\pm$0.005 & 0.9108$\pm$0.006 & \textbf{0.9216}$\pm$0.007\\
PC1 & 0.8470$\pm$0.021 & 0.8468$\pm$0.020 & \textbf{0.8492}$\pm$0.016\\
Tic-Tac-Toe & 0.7334$\pm$0.033 & 0.7314$\pm$0.060 & \textbf{0.7483}$\pm$0.032\\
\midrule
Average & 0.8015$\pm$0.016 & 0.8037$\pm$0.017 & \textbf{0.8090}$\pm$0.013\\
\bottomrule
\end{tabular}
\end{sc}
\end{center}
\vskip -0.1in
\end{table*}

\begin{table*}[!ht]
\label{tab_caafe}
\vskip -0.1in
\caption{Performance comparison between ETPO and CAAFE. CAAFE is the current sota in these datasets that integrates expert-designed models and feature engineering enhanced by GPT-4. From this table, we can notice that ETPO trains a 7B model achieving comparable performance with CAAFE on more than half of tasks}
\vskip 0.05in
\begin{center}
\begin{sc}
\begin{tabular}{l|c|c}
\toprule
Data set & CAAFE(GPT-4) & ETPO(CodeLlama-7B) \\
\midrule
Pharyngitis ⟨\textit{K}⟩ & 0.7078$\pm$0.040 & \textbf{0.7317}$\pm$0.006\\
Kidney Stone ⟨\textit{K}⟩ & \textbf{0.7903}$\pm$0.040 & 0.7881$\pm$0.011\\
Health Insurance ⟨\textit{K}⟩ & 0.5748$\pm$0.020 & \textbf{0.5995}$\pm$0.011\\
Spaceship Titanic ⟨\textit{K}⟩ & 0.8405$\pm$0.020 & \textbf{0.8578}$\pm$0.015\\
\midrule
Airlines & 0.6203$\pm$0.040 & \textbf{0.6580}$\pm$0.011\\
Balance Scale & 0.882$\pm$0.260 & \textbf{0.9289}$\pm$0.038\\
Breast-w & 0.9809$\pm$0.020 & \textbf{0.9948}$\pm$0.003\\
CMC & \textbf{0.7393}$\pm$0.020 & 0.7306$\pm$0.003\\
Credit-g & 0.7832$\pm$0.030 & \textbf{0.7875}$\pm$0.023\\
Diabetes & \textbf{0.8425}$\pm$0.030 & 0.8329$\pm$0.008\\
Eucalyptus & \textbf{0.9319}$\pm$0.000 & 0.8973$\pm$0.006\\
Jungle Chess & \textbf{0.9453}$\pm$0.010 & 0.9216$\pm$0.007\\
PC1 & \textbf{0.9093}$\pm$0.010 & 0.8492$\pm$0.016\\
Tic-Tac-Toe & \textbf{0.9536}$\pm$0.060 & 0.7483$\pm$0.032\\
\bottomrule
\end{tabular}
\end{sc}
\end{center}
\vskip -0.1in
\end{table*}

\clearpage
\section{Ablations}
\label{sec_ablation}

In this section, we conduct ablation studies to investigate the importance of different choices.  

\begin{figure}[!ht]
\label{fig_ablations}
\vskip -0.1in
\begin{center}
\centerline{\includegraphics[width=0.6\linewidth]{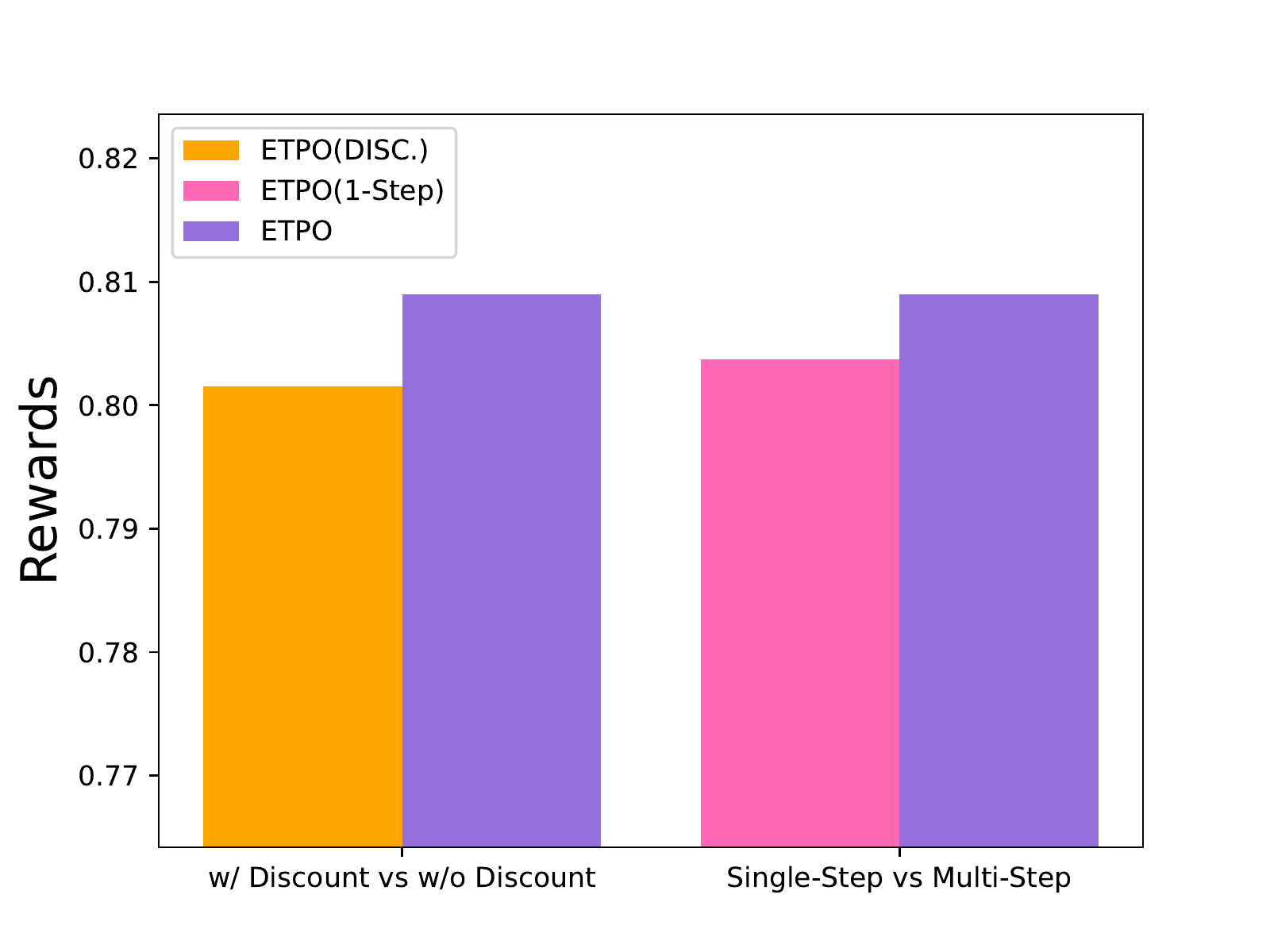}}
\vskip -0.1in
\caption{Average performance comparison between ETPO and its ablated variants, with details in Table~\ref{tab_ablations}. ETPO(DISC.) applies discount factor $\gamma=0.99$ when backpropagating the credit between tokens within an action. ETPO(1-Step) is trained on the environment with a maximum step limit of 1 to disable the reflection process.}
\end{center}
\vskip -0.2in
\end{figure}

\textbf{Discounting within actions or not} In the per-token soft Bellman update, we only discount the Q-values backpropagation between the last token in $a_{t}$ and the first token in the next action $a_{t+1}$. Besides, we also confirm this choice by comparing ETPO with a variant that applies the same discount factor to the credit backpropagation within an action, i.e. the ETPO(DISC.) in Figure~\ref{fig_ablations}. It is worth noting that ETPO(DISC.) is a theoretically flawed algorithm since it breaks the consistency between token optimization and action optimization.

\textbf{Single-step or multi-step}
As we discussed in Section~\ref{sub_sec_convergence}, vanilla CodeLlama-7B is unstable in generating runnable code. To reduce the frequency of generating invalid code and improve exploration efficiency, we model the code generation task as a multi-step sequential decision-making process with reflection. The results in Table~\ref{tab_ablations} confirmed our choice, where ETPO outperforms its single-step variant whose maximum step limit is set to $1$ and the reflection process is ignored.

\end{document}